\documentclass[conference]{IEEEtran}
\IEEEoverridecommandlockouts
\usepackage{cite}
\usepackage{amsmath,amssymb,amsfonts}
\usepackage{algorithmic}
\usepackage{graphicx}
\usepackage{textcomp}
\usepackage{CJKutf8}
\usepackage{listings}
\usepackage{courier}
\usepackage{xcolor}
\usepackage{makecell}
\usepackage{multirow}
\usepackage[hidelinks,bookmarks=false]{hyperref}
\lstdefinestyle{customjson}{
    basicstyle=\ttfamily\scriptsize, 
    numbers=left,
    numberstyle=\tiny,
    stepnumber=1,
    numbersep=5pt,
    showstringspaces=false,
    breaklines=true,
    frame=single,
    backgroundcolor=\color{black!10}, 
    captionpos=b,
    aboveskip=5pt, 
    belowskip=5pt  
}
\def\BibTeX{{\rm B\kern-.05em{\sc i\kern-.025em b}\kern-.08em
    T\kern-.1667em\lower.7ex\hbox{E}\kern-.125emX}}

\hypersetup{
  pdfinfo={
    Title={CREMD: Crowd-Sourced Emotional Multimodal Dogs Dataset},
    Author={},
    Subject={IEEE Conference Submission},
    Keywords={}
  }
}

\begin{document}

\title{CREMD: Crowd-Sourced Emotional Multimodal Dogs Dataset\\
}

\author{\IEEEauthorblockN{1\textsuperscript{st} Jinho Baek}
\IEEEauthorblockA{\textit{dept. of Computer Science} \\
\textit{New York Institute of Technology}\\
New York, USA \\
jbaek01@nyit.edu}
\and
\IEEEauthorblockN{2\textsuperscript{nd} Houwei Cao}
\IEEEauthorblockA{\textit{dept. of Computer Science} \\
\textit{New York Institute of Technology}\\
New York, USA \\
hcao02@nyit.edu}
\and
\IEEEauthorblockN{3\textsuperscript{rd} Kate Blackwell}
\IEEEauthorblockA{\textit{dept. of Computer Science} \\
\textit{New York Institute of Technology}\\
New York, USA \\
kblack01@nyit.edu}
}

\maketitle

\begin{abstract}
Dog emotion recognition plays a crucial role in enhancing human-animal interactions, veterinary care, and the development of automated systems for monitoring canine well-being. However, accurately interpreting dog emotions is challenging due to the subjective nature of emotional assessments and the absence of standardized ground truth methods. We present the CREMD (Crowd-sourced Emotional Multimodal Dogs Dataset), a comprehensive dataset exploring how different presentation modes (e.g., context, audio, video) and annotator characteristics (e.g., dog ownership, gender, professional experience) influence the perception and labeling of dog emotions. The dataset consists of 923 video clips presented in three distinct modes: without context or audio, with context but no audio, and with both context and audio. We analyze annotations from diverse participants—including dog owners, professionals, and individuals with varying demographic backgrounds and experience levels—to identify factors that influence reliable dog emotion recognition. Our findings reveal several key insights: (1) while adding visual context significantly improved annotation agreement, our findings regarding audio cues are inconclusive due to design limitations (specifically, the absence of a no-context-with-audio condition and limited clean audio availability); (2) contrary to expectations, non-owners and male annotators showed higher agreement levels than dog owners and female annotators, respectively, while professionals showed higher agreement levels, aligned with our initial hypothesis; and (3) the presence of audio substantially increased annotators' confidence in identifying specific emotions, particularly anger and fear. 
\end{abstract}


\section{Introduction}
Dogs are highly expressive animals capable of conveying a wide range of emotions through their facial expressions, body language, and vocalizations. Understanding these emotional states is crucial for improving human-animal interactions, veterinary care, and the development of machine learning systems for classifying canine emotions and supporting well-being monitoring. However, accurately interpreting and validating dog emotions remains a significant challenge due to the subjective nature of emotional assessment and the lack of standardized methodologies for establishing ground truth in emotion recognition datasets.

Although prior work has examined aspects of canine emotion recognition, few studies have systematically explored how different modalities and annotator backgrounds influence the annotation process. Many existing datasets rely on annotations from a single source (e.g., general crowdworkers or professionals), or use multi-round strategies without explicitly analyzing how these annotations reflect underlying perceptual reliability or agreement across groups.

In this paper, we present CREMD (Crowd-sourced Emotional Multimodal Dogs Dataset), a comprehensive dataset designed to investigate how different presentation modes and annotator characteristics influence the perception and labeling of dog emotions. Our dataset includes 923 video clips presented in three distinct modes: without context or audio (NCNA), with context but no audio (YCNA), and with both context and audio (YCYA). Through analysis of annotations from diverse participants—including dog owners, professionals, and individuals with varying demographic backgrounds and experience levels—we explore the factors that contribute to consistent and reliable emotion recognition in dogs.

The key contributions of this work include: (1) a novel approach to establishing ground truth through multi-modal presentation and diverse annotator groups; (2) comprehensive analysis of how contextual information and audio cues affect both annotation agreement and annotator confidence; (3) investigation of demographic factors influencing agreement patterns across rater groups; and (4) insights into the challenges and opportunities in developing machine learning models for automated dog emotion recognition.

\section{Related Works}
This section discusses the current state of research in canine emotion recognition, focusing on three key areas: visual recognition, audio analysis, and multimodal approaches.

\subsection{Visual Recognition of Dog Emotions}
Recent studies have made significant progress in understanding and detecting dog emotions through visual cues. Boneh-Shitrit et al. [4] developed deep learning models for automated classification of dog emotional states from facial expressions, achieving promising results in distinguishing between different emotional states. This work was extended in their subsequent research [5], which focused on recognizing specific emotional states like positive anticipation and frustration. Chavez-Guerrero et al. [7] employed computer vision techniques for classifying domestic dogs' emotional behavior, while Chen et al. [8] proposed a convolutional neural network-based system for automated dog tracking and emotion recognition in video surveillance. These studies demonstrate the growing capability of machine learning systems to detect and classify canine emotions through visual analysis. The importance of facial expressions in dog emotion recognition was further explored by Waller et al. [22], who investigated how paedomorphic facial expressions give dogs a selective advantage in human interactions. This research was complemented by Schirmer et al. [21], who found that humans process dog and human facial affect in similar ways, suggesting shared mechanisms in emotional recognition across species.

\subsection{Audio-Based Emotion Analysis}
Several researchers have focused on analyzing dog vocalizations to understand emotional states. Abzaliev et al. [1] explored leveraging human speech processing techniques for automated bark classification. Hantke et al. [13] developed systems for automatic recognition of context and perceived emotion in dog barks, while Maskeliunas et al. [17] conducted comprehensive studies on recognizing emotional vocalizations of canines. Pongrácz et al. [20] demonstrated that acoustic parameters of dog barks carry emotional information that humans can interpret, providing a foundation for audio-based emotion recognition systems. Perez-Espinosa et al. [19] advanced this field by applying machine learning approaches to analyze vocalizations using a bidimensional model of emotions.

\subsection{Multimodal and Comprehensive Approaches}
Recent research has increasingly focused on combining multiple modalities for more accurate emotion recognition. Correia-Caeiro et al. [9] provided a critical review of methodologies in visual perception of emotion cues in dogs, while Csoltova and Mehinagic [10] conducted a state-of-the-art review of positive-emotion assessment in domestic dogs. The ability of dogs themselves to recognize emotions has been studied by Albuquerque et al. [2], who found that dogs can recognize both dog and human emotions. Cultural aspects of emotion recognition were explored by Amici et al. [3], who discovered that the ability to recognize dog emotions depends on cultural background. Recent technological advances have enabled more sophisticated approaches to emotion recognition. Ferres et al. [11] used DeepLabCut for posture analysis in predicting dog emotions, while Franzoni et al. [12] conducted preliminary work on comprehensive dog emotion recognition systems. Kowalczuk et al. [15] employed deep neural networks for categorizing emotions in dog behavior. Bremhorst et al. [6] conducted detailed studies on differences in facial expressions during positive anticipation and frustration in dogs awaiting rewards, while Hernández-Luquin et al. [14] developed the DEBIw dataset for recognizing dog emotions from images in the wild, contributing to the growing body of research in this field. Mota-Rojas et al. [18] reviewed current advances in assessing dogs' emotions and facial expressions, particularly focusing on clinical pain recognition. Kujala [16] provided essential guidelines for research in canine emotions, emphasizing the importance of standardized methodologies and comprehensive approaches. These diverse research approaches have contributed to a growing understanding of how dogs express emotions and how these expressions can be accurately detected and classified through various technological means. The field continues to evolve with new methodologies and technologies, moving towards more accurate and comprehensive emotion recognition systems.

\section{Data Collection}

\begin{table}[htbp]
\caption{Keywords for Dog Emotion Categories}
\begin{CJK*}{UTF8}{mj}
\begin{tabular}{|l|l|l|l|}
\hline
\textbf{Fear} & \textbf{Happy} & \textbf{Angry} & \textbf{Neutral} \\
\hline
scared & playful & Agressive & calm \\
rescue & toy & fight & relaxed \\
abandoned & friendly & food agression & rest \\
shelter & excited & resource guarding & stting quietly \\
anxious & play & cesar milan & neutral state \\
anxiety & ball & growl & no emotion \\
stray dog & joyful & bit & lying down \\
concerned & owner & bark & passive \\
terrified & zoomies & Perro enojado & chill \\
stressed & fetch & growling & in bed \\
Separation anxiety & fun & pit bull & tired \\
nervous & Perro feliz & bull terrier & bored \\
frightened & open mouth & bitten & sleepy \\
worried & wag tail & territorial &  \\
thunder & &  & \\
trauma & &  & \\
\hline
\end{tabular}
\end{CJK*}
\label{tab:dog_emotion_keyword}
\end{table}

\begin{table}[htbp]
\caption{Distribution of Dog Appearances Across Videos}
\begin{center}
\begin{tabular}{|c|c|c|c|c|c|c|c|}
\hline
\textbf{Appearances} & \textbf{1} & \textbf{2} & \textbf{3} & \textbf{4} & \textbf{5} & \textbf{6} & \textbf{Total} \\
\hline
Count & 313 & 31 & 4 & 0 & 2 & 1 & 351 \\
\hline
Percentage & 89\% & 9\% & 1\% & 0\% & 1\% & 0\% & 100\% \\
\hline
\end{tabular}
\label{tab:dog_appearances}
\end{center}
\end{table}

\begin{table*}[!htbp]
\caption{Example entries from the dog metadata database}
\begin{center}
\begin{tabular}{|c|c|c|c|c|c|c|c|c|}
\hline
\textbf{Index} & \textbf{Dog Id} & \textbf{Dog Name} & \textbf{Dog Breed} & \textbf{Dog Age} & \textbf{Dog Color} & \textbf{Dog Gender} & \textbf{Dog Size} & \textbf{Videos} \\
\hline
1 & D1003 & Tori & - & - & golden/white & - & medium & V1003 \\
2 & D1056 & Dory & - & - & tan/light brown & female & big & V1058, V1070, V1071 \\
3 & D1205 & Cong & miniature poodle & - & light brown & male & small & V1237, V1238 \\
4 & D1146 & Meira & pit bull terrier & - & grey/white & female & medium & V1161, V1380, V1381 \\
\hline
\end{tabular}
\label{tab:dog_prelim_metadata}
\end{center}
\end{table*}

Based on a review of more than 20 articles on canine emotion detection [1-22], we identified the categories of canine emotion that are most frequently studied: happy, angry, fearful, and neutral. To facilitate comprehensive video searches on platforms such as YouTube, Reels, TikTok, and others, we developed a list of relevant keywords for each emotion class, as shown in Table \ref{tab:dog_emotion_keyword}. Using these keywords, we performed video searches on YouTube, Instagram, TikTok, and Pexels, collecting approximately 100 videos for each preliminary class category, totaling 440 videos. After reviewing each video, we created a comprehensive spreadsheet of dog metadata. The dataset contains 351 unique dogs, most of which appear in a single video. As shown in Table \ref{tab:dog_appearances}, a small number of dogs appear multiple times, with one dog featured in six videos, the maximum number of appearances. Table \ref{tab:dog_prelim_metadata} presents examples of the detailed dog metadata we collected.

\subsection{Data Cleaning \& Prepossessing}

\begin{figure}[h]
\centering
\includegraphics[scale=0.15]{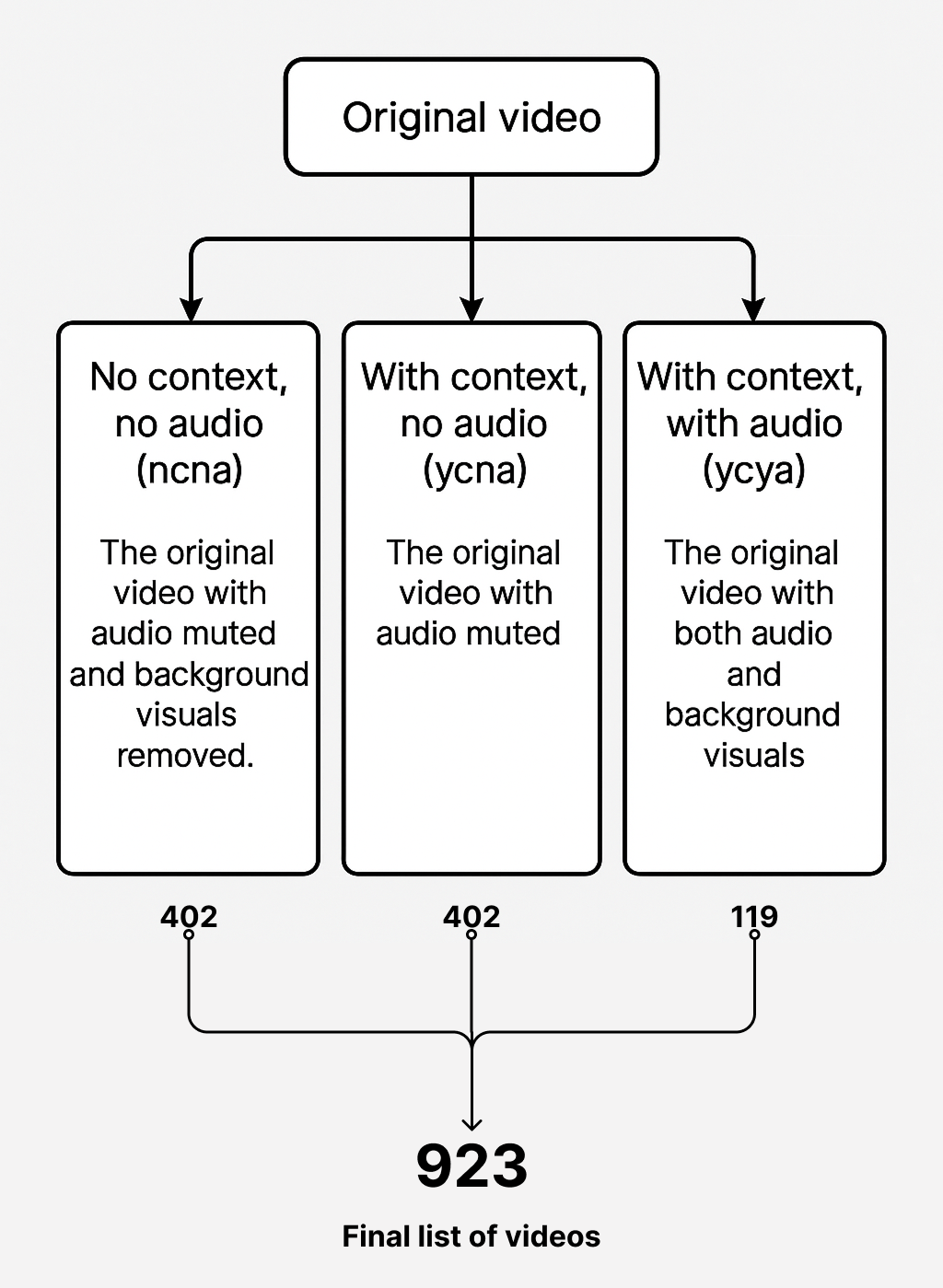}
\caption{Three versions of the video clips created for dog emotion annotation}
\label{fig:distribution_dataset}
\end{figure}

We reviewed all 440 videos and removed those with poor visual quality, partial segments, shortened versions, and duplicate recordings from our dataset. After cleaning, 402 videos remained. In addition, comprehensive metadata were created to provide detailed information about each video, including context, environment, audio presence, and quality. 
Next, to examine how video context and audio information affect human perception of dog emotions, we created three versions of the data clips for dog emotion annotation, as shown in Fig. \ref{fig:distribution_dataset}.

\begin{itemize} 
\item \textbf{With context, with audio (YCYA)}: The original video with both audio and visual context. This version tests whether the combination of visual context and audio cues enhances annotators' ability to predict dog emotions. 
\item \textbf{With context, no audio (YCNA)}: The original video with muted audio and visual context. This version tests whether background visuals alone, without audio, improve annotators' ability to predict dog emotions. 
\item \textbf{No context, no audio (NCNA)}: The video with both muted audio and removed context. We used a pre-trained YOLOv8m model to detect and crop only the portion containing the dog. \end{itemize}

We note that a no-context-with-audio (NCYA) condition was not included due to two key constraints. First, the majority of publicly available videos with audio did not contain discernible dog-generated vocalizations. When such sounds were present, they were typically limited to certain emotional states like fear or anger, and often absent, inaudible, or ambiguous in others such as happy or neutral. Second, even when audio was present, it was rarely clean, as most videos included overlapping background noise—such as human speech or music—making it difficult to isolate dog-generated sounds. As a result, the limited availability of clean audio prevented inclusion of a robust NCYA condition and also led to a smaller YCYA subset.   

Finally, we have a total of 923 videos for dog emotion annotation: 402 in NCNA, 402 in YCNA, and 119 in YCYA, with an average duration of 15.8 seconds per video. 

\vspace{10pt}  
\section{Dog Emotion Annotation}

Our primary goals for this study were threefold. First, we aimed to collect more than 12 ratings for each of the 923 videos. Second, we sought to investigate how video context and audio influence human perception of dog emotions. Third, we aimed to understand whether specific demographic factors, such as dog ownership experience, professional experience, and gender, influenced the annotation of dog emotions.

\begin{figure}[htbp]
\centerline{\includegraphics[scale=0.20]{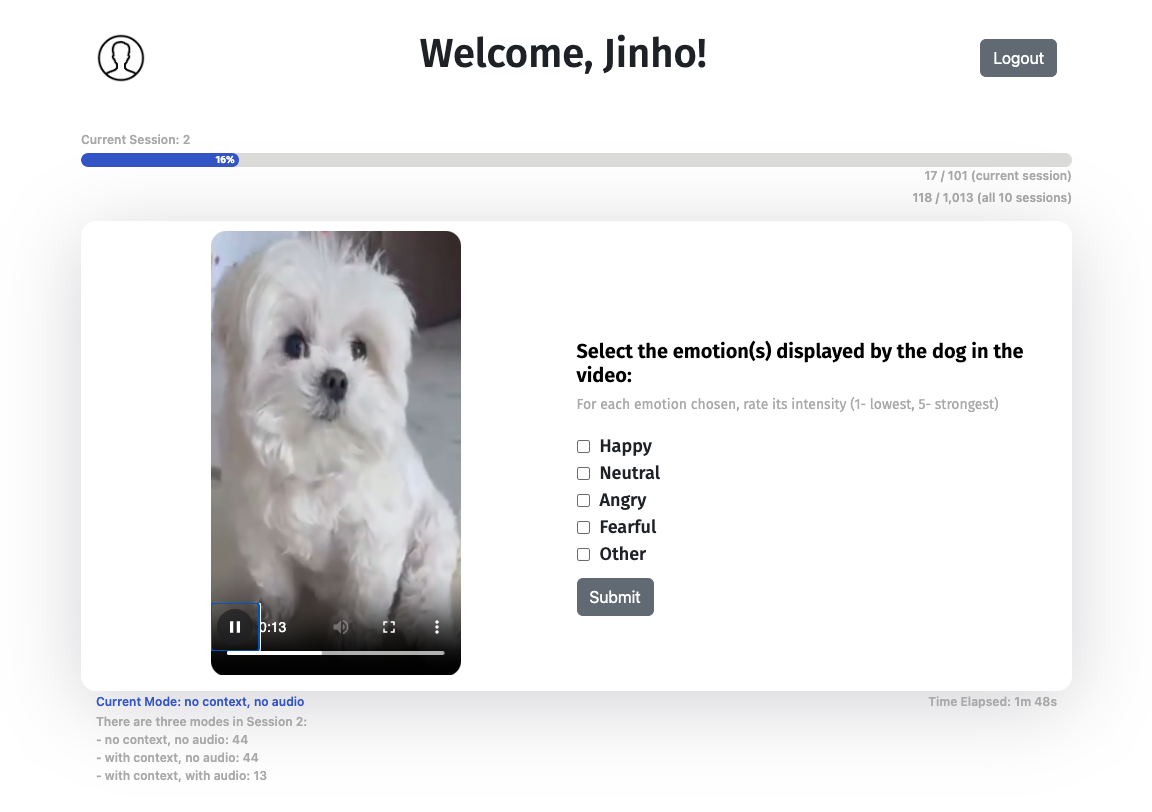}}
\caption{Interface of the custom-developed video annotation tool used for dog emotion labeling}
\label{fig:annotation_tool}
\end{figure}

After evaluating platforms like Amazon Mechanical Turk and alternative tools like Roboflow, Anvil, and Annemo, we developed a custom annotation tool. While platforms like Amazon Mechanical Turk offer convenience, we had limited control over task design and quality assurance. Our custom tool allowed us to tailor the interface, enforce stricter quality checks, and collect the specific data we needed for our research objectives.
As shown in Fig. \ref{fig:annotation_tool}, we created this tool using React.js for the frontend and Firebase for the backend and database management. The videos are organized into 10 sessions, each containing approximately 100 clips, presented in a specific order with randomized conditions to prevent bias: 40 clips without context/audio, 40 clips with context but without audio, and 12 clips with both context and audio. To ensure consistency, 10\% of the videos are repeated for each annotator. The full annotation task consisted of 1,013 videos split into 10 sessions, designed to take approximately 3 hours total. Feedback and response-time analysis suggest many annotators spent 4–5 hours. To mitigate fatigue, annotators were given one week to complete the task, with guidance to do one session per sitting (or up to three per day), and the option to pause and resume mid-session.

Each annotator first completes a demographic survey before receiving a unique sequence of videos to annotate. During the data annotation process, annotators watch each video and select one or more emotion labels. For each emotion, they rate its intensity on a scale of 1 to 5, with 5 indicating the highest confidence (default rating is 3). If the standard emotion categories do not capture their observations, annotators can select "Other" and provide a written description. The system tracks annotator progress and response times, while allowing video replays as needed.

A total of 23 annotators were recruited to participate in the dog emotion annotation task, and their demographic breakdown is shown in Table IV. We acknowledge that this sample size is relatively small for drawing definitive conclusions; however, several factors mitigate this concern. First, depth of annotation: 10 out of 23 annotators completed the full set of 1,013 videos, and as shown in Fig. \ref{fig:annotation_count_histogram}, over 93\% of clips have more than 8 annotations, with the majority receiving 14 or more. Second, diversity over quantity: due to environmental constraints, we prioritized annotator diversity over volume. Our sample reflects a broad range of demographic backgrounds and lived experiences with dogs. The group included slightly more females than males, with a 13:10 female-to-male ratio. The majority were dog owners, comprising 16 owners and 7 non-owners. Among the participants, 8 were professionals in the field, while 15 were non-professionals. The median age was 32 years. In terms of racial diversity, the group consisted of 13 Asians, 6 Caucasians, 3 Hispanics, and 1 individual who identified as another race. Participants were recruited through targeted outreach to ensure a diverse range of backgrounds, including both pet owners and individuals with professional experience in animal care. That said, we acknowledge that a larger and more demographically balanced sample would improve the robustness of subgroup analyses, and expanding the annotator pool is a key priority in future iterations of the dataset.

\begin{table}[htbp]
\caption{Demographic breakdown of study participants}
\begin{center}
\begin{tabular}{|l|l|}
\hline
\textbf{Category} & \textbf{Statistic} \\
\hline
Size & 23 \\
Female : Male & 13:10 \\
Owners & 16 (70\%) \\
Professionals & 8 (35\%) \\
Median Age & 32 \\
Race (n) & Asian (13), Caucasian (6), Hispanic (3), Other (1) \\
\hline
\end{tabular}
\label{tab:study_participants}
\end{center}
\end{table}

\begin{table}[htbp]
\caption{Total annotations and votes in the dataset}
\begin{center}
\begin{tabular}{|c|c|}
\hline
\textbf{Category} & \textbf{Count} \\
\hline
Annotations & 12,794 \\
Votes & 15,195 \\
\hline
\end{tabular}
\label{tab:annotations_votes}
\end{center}
\end{table}

\begin{table}[htbp]
\caption{Distribution of Votes per Annotation Event}
\begin{center}
\begin{tabular}{|c|c|c|}
\hline
\textbf{Votes per Annotation} & \textbf{Count} & \textbf{\%} \\
\hline
3 & 70 & 1\% \\
2 & 2,261 & 18\% \\
1 & 10,463 & 82\% \\
\hline
Total & 12,794 & 100\% \\
\hline
\end{tabular}
\label{tab:votes_distribution}
\end{center}
\end{table}

\section{Human Perception of Dog Emotion}

To better understand how people perceive dog emotional expressions across different modalities, we examine response times, inter-rater agreement and consistency, label intensities, and modality-based annotation patterns.

\subsection{Distribution of Annotations and Vote Counts}

As shown in Table \ref{tab:annotations_votes}, our data collection process yielded a comprehensive dataset of 12,794 distinct annotations and 15,195 total votes across 923 preprocessed video clips. The higher number of votes compared to annotations indicates that annotators frequently identified multiple emotional expressions per video, reflecting the complexity of emotional content in the clips, as further illustrated in Table \ref{tab:votes_distribution}.

\begin{figure}[h]
\centering
\includegraphics[scale=0.5]{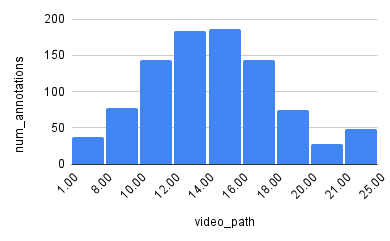}
\caption{Distribution of annotation counts showing normal curve across 923 videos with 12,794 total annotations. The x-axis represents video identifiers (video\_id), and the y-axis shows the number of annotations per video.}
\label{fig:annotation_count_histogram}
\end{figure}

The distribution of these annotations, shown in Fig. \ref{fig:annotation_count_histogram}, reveals that over 95\% of videos received at least 8 annotations. This non-uniform distribution resembles a normal curve and likely emerged due to two factors: the inclusion of repeat videos, which increased annotation counts for certain clips, and uneven participation in the crowd-sourced setting, where some annotators completed only one or two sessions.

\subsection{Response Time}

\begin{figure}[h]
\centering
\includegraphics[scale=0.3]{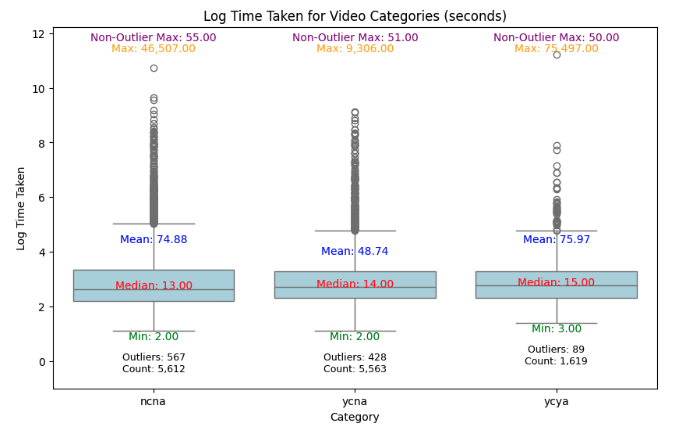}
\caption{Box plots comparing response times across three video modes}
\label{fig:response_times_all}
\end{figure}

The videos in our dataset averaged 15.8 seconds in length, with NCNA clips slightly shorter due to pre-processing. Fig. \ref{fig:response_times_all} compares the distribution of response times across modalities. The median response times were 13, 14, and 15 seconds for NCNA, YCNA, and YCYA, respectively—closely matching average video durations. This suggests that annotators typically responded shortly after watching each clip.

The mean response times exhibited a right skew due to outliers, likely caused by breaks taken between sessions. Still, the data suggest strong engagement, as annotators often re-watched clips and adjusted their responses.

As more context was provided, mean response times decreased from 74.88 seconds (NCNA) to 48.74 seconds (YCNA), reflecting a 35\% reduction. This suggests that context helped facilitate quicker decisions. Although YCYA showed longer response times overall, this was likely due to a subset of clips with longer durations. The percentage of extreme response times decreased as context and audio were added, suggesting greater annotator confidence.

\subsection{Inter-Rater Agreement and Consistency}

\begin{figure}[h]
\centering
\includegraphics[scale=0.3]{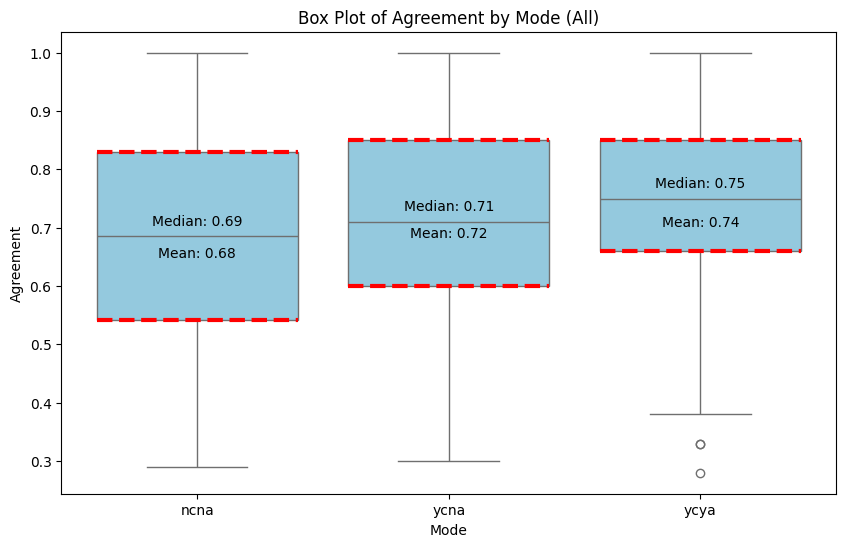}
\caption{Box plots showing annotation agreement rates (0.0–1.0) across the three video presentation modes for all videos}
\label{fig:agreement_all}
\end{figure}

The agreement level for each video is defined as the ratio of the most frequent emotion vote to the total number of votes. As shown in Fig. \ref{fig:agreement_all}, videos with more contextual and auditory information exhibited higher median agreement values: increasing from 0.69 in NCNA to 0.71 in YCNA, and to 0.75 in YCYA, with a smaller spread across users.

The Shapiro-Wilk test results (p-values ${<}$ 0.05) indicated that agreement scores across all three modalities) indicated that agreement scores were not normally distributed. We note that the Shapiro-Wilk test is used to assess the assumption of normality for parametric tests, and it is known to be sensitive to sample size—with large sample sizes, the test is likely to be significant even for minor deviations from normality. Given the non-normal distributions, we used the non-parametric Kruskal-Wallis H-test, which confirmed significant differences in agreement values (p $\approx$ 0.0025). This suggests that the amount of information provided significantly affects annotation consistency.

We further conducted pairwise comparisons using the Mann-Whitney U test (Table \ref{tab:pairwise_comparisons_agreement}). While the overall Kruskal-Wallis test showed a significant difference, pairwise results revealed that NCNA differed significantly from both YCNA (p = 0.008) and YCYA (p = 0.003), whereas YCNA and YCYA did not significantly differ (p = 0.173). This indicates that adding visual context improves agreement. However, our findings regarding audio's impact on agreement are inconclusive, as we were unable to include a no-context-with-audio condition and had limited clean audio availability, resulting in a smaller YCYA subset. These results suggest that visual context helps standardize annotator interpretation across raters, while the role of audio cues requires further investigation with balanced experimental conditions.

As a quality control measure, we evaluated annotator self-consistency on repeated videos. Annotator consistency is defined as the proportion of repeated clips for which the same label was assigned across all views. Complete consistency indicates identical annotations across repeats, partial consistency denotes overlap in at least one emotion label, and complete inconsistency indicates no overlap. We note that different ratings (e.g., intensity ratings of 1 and 5) should not be treated equally when interpreting consistency, and that interpreting rating variability is nontrivial. To support a range of analytical approaches, the dataset includes both individual- and group-level annotations for perceived emotions and intensities, allowing researchers to re-weight or segment responses according to their specific goals. Rather than enforcing a single aggregation method, the design intentionally preserves variability across modalities and annotator subgroups.

Table \ref{tab:repeat_consistency} confirms high self-consistency: 68\% of annotations were completely consistent, 27\% partially consistent, and only 5\% fully inconsistent. As more context and audio were added, inconsistency decreased, and total consistency increased.

\begin{table}[!htbp]
\caption{Pairwise Comparison Results of Agreement}
\begin{center}
\begin{tabular}{|c|c|}
\hline
\textbf{Comparison} & \textbf{p-value} \\
\hline
NCNA vs. YCNA & 0.008 \\
YCNA vs. YCYA & 0.173 \\
NCNA vs. YCYA & 0.003 \\
\hline
\end{tabular}
\label{tab:pairwise_comparisons_agreement}
\end{center}
\end{table}

\begin{table}[!htbp]
\caption{Average inter-annotator agreement scores}
\begin{center}
\begin{tabular}{|c|c|}
\hline
\textbf{Description} & \textbf{Value Range} \\
\hline
Agreement scores & 0.0 to 1.0 \\
\hline
0.0 & No agreement above chance \\
1.0 & Perfect agreement \\
\hline
\end{tabular}
\label{tab:agreement_scores_explanation}
\end{center}
\end{table}

\begin{table}[htbp]
\caption{Distribution of annotation consistency across repeat videos}
\begin{center}
\begin{tabular}{|l|c|c|c|c|}
\hline
\textbf{Category} & \textbf{NCNA} & \textbf{YCNA} & \textbf{YCYA} & \textbf{Overall} \\
\hline
Not consistent & 3\% & 2\% & 0\% & 5\% \\
Partly consistent & 13\% & 11\% & 3\% & 27\% \\
Consistent & 29\% & 30\% & 9\% & 68\% \\
\hline
\end{tabular}
\label{tab:repeat_consistency}
\end{center}
\end{table}

\subsection{Dog Emotion Perception in Various Modalities}

\begin{table*}[htbp]
\caption{Statistical measures across demographic groups}
\begin{center}
\begin{tabular}{|l|c|c|c|c|c|c|}
\hline
\textbf{Statistic} & \textbf{Male} & \textbf{Female} & \textbf{Owner} & \textbf{Non-owner} & \textbf{Professional} & \textbf{Non-professional} \\
\hline
median & \textbf{0.79} & 0.67 & 0.69 &\textbf{0.80} & \textbf{0.75} & 0.70 \\
mean & 0.76 & 0.69 & 0.69 & 0.78 & 0.76 & 0.71 \\
1QR & 0.60 & 0.56 & 0.56 & 0.60 & 0.60 & 0.57 \\
3QR & 1.00 & 0.83 & 0.83 & 1.00 & 1.00 & 0.83 \\
max (excl. outliers) & 1.00 & 1.00 & 1.00 & 1.00 & 1.00 & 1.00 \\
min (excl. outliers) & 0.25 & 0.25 & 0.28 & 0.25 & 0.25 & 0.25 \\
IQR & 0.40 & 0.28 & 0.28 & 0.40 & 0.40 & 0.26 \\
\hline
\end{tabular}
\label{tab:agreement_by_demographics}
\end{center}
\end{table*}

\begin{figure}[h]
\centering
\includegraphics[scale=0.6]{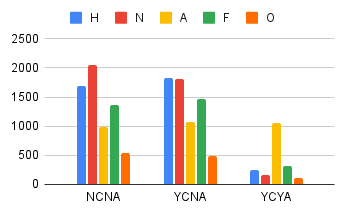}
\caption{Distribution of emotion labels across three modalities}
\label{fig:emotion_distr_column}
\end{figure}

The distribution of emotion labels across modalities, shown in Fig. \ref{fig:emotion_distr_column}, reveals clear trends. In NCNA videos, Neutral received the most votes, likely due to the limited information available. When visual context was added (YCNA), votes shifted from Neutral toward Happy, with slight increases in Fear and Anger, and a decrease in Other selections, suggesting more confident interpretations and that contextual information helped annotators assign more specific emotion labels rather than defaulting to ambiguous or neutral responses. YCYA annotations were dominated by Anger, likely due to the presence of distinctive audio cues such as growling. Videos containing other emotions tend to lack similarly specific auditory signals. This result may also reflect the composition of the YCYA set, which included only 30\% of the 402 base videos and disproportionately featured emotionally intense clips.

\begin{figure}[h]
\centering
\begin{minipage}{0.8\linewidth}
\includegraphics[scale=0.3]{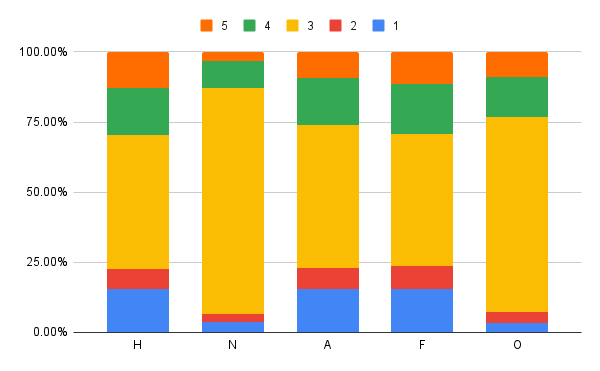}
\end{minipage}
\begin{minipage}{0.8\linewidth}
\includegraphics[scale=0.3]{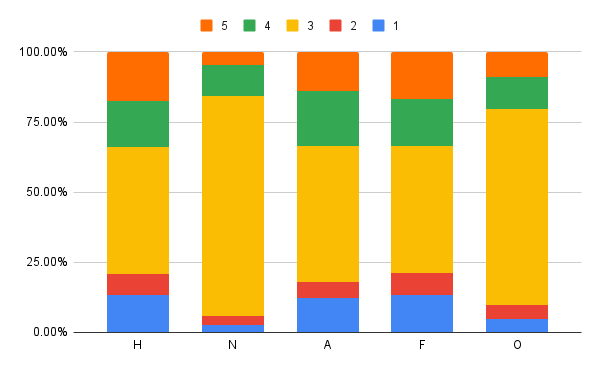}

\end{minipage}
\begin{minipage}{0.8\linewidth}
\includegraphics[scale=0.3]{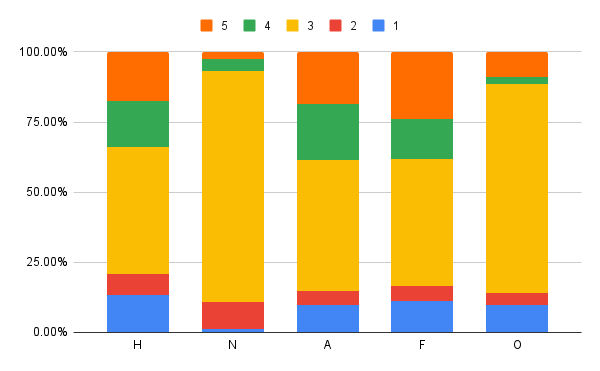}
\caption{The distribution of emotion intensity ratings (as percentages) across the three modalities: NCNA (top), YCNA (middle), and YCYA (bottom).}
\label{fig7:c}
\end{minipage}
\end{figure}

Fig. \ref{fig7:c} shows the distribution of intensity ratings across modalities. For NCNA clips, annotators most often selected intensity level 3 for Neutral and Other. For more expressive emotions, they tended toward extremes—level 1 for low and level 4 for high intensity. In YCNA, intensity level 5 became more frequent for distinct emotions like Happy, Anger, and Fear. This pattern was further amplified in YCYA, where level 5 ratings for Anger and Fear were more common. These findings suggest that context and audio together enhance annotator confidence in identifying intense emotions.

We also observed notable ambiguity in emotion interpretation. In videos without audio, Happy was often confused with Neutral, and Fear with Other. With context and audio, the most frequent confusion occurred between Anger and Fear, which sometimes share vocalizations (e.g., growling) and physical cues (e.g., tense posture). This highlights the interpretive complexity of canine emotions, even when multiple sensory cues are present.

\vspace{10pt}  
\section{Dog Emotion Perception across Demographics}

We now turn to dog emotion perception across different demographic groups and experiential backgrounds, including gender, dog ownership, and professional experience. As shown in Table \ref{tab:agreement_by_demographics}, non-owners and male annotators consistently demonstrated higher levels of agreement than dog owners and female annotators, respectively. This suggests that non-owners and males may rely more heavily on observable cues such as body language or facial expressions, while dog owners and females could be more attuned to subtle emotional nuances, leading to a greater diversity of interpretations and thus slightly lower agreement.

Similarly, we observe that professional annotators—those engaged in fields such as training, dog grooming, or dog walking—achieved higher agreement levels than non-professionals, which aligns with our initial hypothesis. This supports the idea that professionals, by virtue of their extensive and varied exposure to dogs, develop a more objective understanding of canine behavior. Their experience-based perspective allows them to interpret ambiguous or subtle emotional cues with greater consistency across instances.

\subsection{Statistical Analysis}

\begin{table}[htbp]
\caption{Statistical Test Results: Owner vs. Non-owner Comparisons}
\begin{center}
\begin{tabular}{|c|c|c|}
\hline
\textbf{Test Type} & \textbf{Group} & \textbf{Value} \\
\hline
Shapiro Test (Owners) & p-value & 1.72 × 10\textsuperscript{-13} \\
Shapiro Test (Non-owners) & p-value & 1.28 × 10\textsuperscript{-25} \\
Mann-Whitney U Test & p-value & 1.18 × 10\textsuperscript{-21} \\
\hline
\end{tabular}
\label{tab:owner_stats}
\end{center}
\end{table}

\begin{table}[htbp]
\caption{Statistical Test Results: Male vs. Female Comparisons}
\begin{center}
\begin{tabular}{|c|c|c|}
\hline
\textbf{Test Type} & \textbf{Group} & \textbf{Value} \\
\hline
Shapiro Test (Male) & p-value & 1.95 × 10\textsuperscript{-21} \\
Shapiro Test (Female) & p-value & 4.14 × 10\textsuperscript{-15} \\
Mann-Whitney U Test & p-value & 6.21 × 10\textsuperscript{-15} \\
\hline
\end{tabular}
\label{tab:gender_stats}
\end{center}
\end{table}

\begin{table}[htbp]
\caption{Statistical Test Results: Professional vs. Non-professional Comparisons}
\begin{center}
\begin{tabular}{|c|c|c|}
\hline
\textbf{Test Type} & \textbf{Group} & \textbf{Value} \\
\hline
Shapiro Test (Professionals) & p-value & 2.70 × 10\textsuperscript{-22} \\
Shapiro Test (Non-professionals) & p-value & 2.50 × 10\textsuperscript{-14} \\
Mann-Whitney U Test & p-value & 5.48 × 10\textsuperscript{-8} \\
\hline
\end{tabular}
\label{tab:prof_stats}
\end{center}
\end{table}

\begin{itemize}

\item \textbf{Owners vs. Non-owners}: 
As shown in Table \ref{tab:owner_stats}, statistical analysis revealed significant differences between dog owners and non-owners. The Shapiro test p-values for both groups (1.28e-25 for non-owners and 1.72e-13 for owners) were well below 0.05, indicating non-normal distributions. Given this non-normality, we used the Mann-Whitney U test, which showed a highly significant difference between the groups (p = 1.18e-21). These results strongly support the observed counterintuitive agreement pattern, where individuals without dog ownership experience displayed more consistent emotion annotations.

\item \textbf{Male vs. Female}: 
As shown in Table \ref{tab:gender_stats}, our analysis examined differences between male and female annotators. The Shapiro test revealed non-normal distributions for both groups, with p-values well below 0.05 (1.95e-21 for males and 4.14e-15 for females). Based on this non-normality, we conducted a Mann-Whitney U test, which revealed a highly significant difference between male and female annotators (p = 6.21e-15). These findings suggest that male annotators tended to reach more consistent agreement, potentially due to a greater reliance on overt emotional signals, in contrast to females who may attend to more layered or context-dependent emotional expressions.

\item  \textbf{Professionals vs. Non-professionals}: 
Statistical analysis of Table \ref{tab:prof_stats} revealed non-normal distributions for both professional and non-professional groups, with Shapiro test p-values below 0.05. A Mann-Whitney U test indicated a highly significant difference between the two groups (p $\approx$ 5.48e-08). These results further support the distinct annotation consistency observed among professionals. Their training and practical experiences likely offer them a broader internal reference set for interpreting canine emotional cues, resulting in more aligned interpretations within their cohort.
\end{itemize}

\subsection{Discussion}
\begin{itemize}

\item \textbf{Limitations in Demographic Comparisons:} We acknowledge that comparing owners/non-owners and male/female groups only makes sense if groups are balanced, and overlapping demographic variables may confound subgroup comparisons. While we made efforts to minimize imbalance, complete stratification across all relevant variables (e.g., ownership, profession, gender) was not feasible in this iteration. In future work, we plan to adopt more rigorous methods—such as matched sampling, stratification, or regression-based controls—to ensure greater balance and improve the interpretability of group-level comparisons.

\item \textbf{Higher Consistency and Lower Variability:} 

Higher consistency in emotion ratings does not necessarily reflect superior emotional perceptiveness. In fact, it may point to a more reductive or less nuanced interpretation of emotional states. Female annotators, who are often characterized by heightened emotional sensitivity, may be capable of perceiving multiple emotional states within a single video. This perceptual richness can increase variability in responses, as individuals may weigh cues differently. In contrast, male annotators may attend more to dominant or overt signals, leading to more consistent—but potentially less sensitive—emotion labeling.

\item \textbf{Dog Owners' Biases:} 
Dog owners may carry implicit biases shaped by their personal experiences with their own pets. For instance, when viewing a dog in a neutral or ambiguous pose, an owner may project prior emotional associations tied to similar expressions in their own dog. These emotionally colored memories could influence labeling behavior, introducing subjectivity and resulting in less agreement across the group. Such familiarity, while enriching in some ways, may introduce interpretive variability not present in more detached observers.

\item \textbf{Professional Annotators:} 
Professionals—such as trainers, groomers, or dog walkers—tend to show high agreement levels across modalities. Their occupational exposure to a wide array of dog behaviors enables them to build a more objective internal model of canine emotion expression. This broader base of observational experience helps filter out personal biases and supports higher annotation consistency. Their familiarity with emotional signals in a variety of contexts likely enhances their ability to interpret subtler cues that less experienced annotators may overlook.
\item \textbf{Theoretical Insights:} 

One promising approach to improving emotion annotation quality may lie in combining multiple annotator profiles. By integrating the intuitive, emotionally sensitive perspectives of female annotators and dog owners with the structured, objective judgments of professionals, future annotation systems could achieve a more holistic and balanced understanding of canine emotion. Such a hybrid annotation strategy could capture both nuanced emotional dynamics and generalized consensus, yielding more ecologically valid datasets.

\item \textbf{Gender Comparison:} 

Gender differences in agreement may, in part, be confounded by the overlap between gender and ownership experience. In our sample, 77\% of female annotators were also dog owners, suggesting that perceived gender effects might actually reflect owner-specific patterns. Conversely, male annotators had a more balanced distribution between owners and non-owners, enabling clearer interpretation of gender-specific trends. This demographic overlap highlights the need to disentangle intersecting identity factors when analyzing perception and labeling behavior.
\end{itemize}

\vspace{10pt}  

\section{Conclusion}
In conclusion, this study highlights the complexities and challenges involved in recognizing canine emotions, particularly in the context of crowd-sourced annotation. The CREMD dataset offers insights into how modalities—such as visual context and audio—impact emotion perception, and how factors like dog ownership, gender, and professional experience affect annotation agreement. Our findings suggest that while visual context significantly enhances annotation consistency, our findings regarding audio's impact on agreement are inconclusive due to design limitations (specifically, the absence of a no-context-with-audio condition and limited clean audio availability). However, audio cues do appear to improve annotator confidence in identifying specific emotions, particularly anger and fear.

The demographic analysis revealed counterintuitive patterns, with non-owners and male annotators showing higher agreement levels than their counterparts, while professionals consistently demonstrated the highest agreement across conditions, in line with our initial expectations. These findings suggest that background experience and gender may influence emotional perception in more complex ways than previously assumed.

Overall, this research underscores the need for continued exploration and refinement in both human and machine approaches to canine emotion recognition. The CREMD dataset serves as a foundational resource to support future studies, including both behavioral research and computational modeling of dog emotions. The dataset will be made available to the academic community for research purposes upon acceptance, allowing researchers to apply their preferred methodologies tailored to specific goals. We hope this resource will advance the field of canine emotion recognition and promote the development of more accurate and insightful tools for understanding dog emotions.

\newpage

\section*{Ethical Impact Statement}

This work introduces the Crowd-sourced Emotional Multimodal Dogs Dataset (CREMD), developed to advance canine emotion recognition through multimodal data and diverse annotator perspectives. Given the nature of our research, we identify and address several ethical considerations relevant to human subject involvement, potential societal impact, and limits of generalizability.

\textbf{Human Subjects and Consent:} Annotation data was collected from 23 individuals who voluntarily participated after providing informed consent. Participants completed a demographic survey and were informed of the study’s goals, data usage policies, and their right to withdraw. The study did not involve any medical, psychological, or physical intervention. An ethics oversight process was followed at our institution, and data collection adhered to IRB-equivalent standards. No personally identifiable information (PII) was collected or stored beyond anonymized demographics necessary for analysis. Data were managed using secure systems with access restricted to authorized researchers.

\textbf{Animal Welfare Considerations:} All videos were sourced from public platforms (YouTube, TikTok, Instagram, Pexels), and no animals were harmed or induced into emotional states for the purpose of this study. Video selection prioritized ethical considerations, avoiding content that depicted animal distress beyond everyday interactions. No direct interaction with animals occurred during the dataset creation.

\textbf{Bias and Generalizability:} Our findings highlight demographic-related variability in annotation agreement (e.g., higher consistency among male and non-owner annotators). While this provides valuable insight into human perception, it also raises concerns regarding potential bias if models are trained without diverse perspectives. Annotator background (e.g., dog ownership, gender, professional experience) significantly influenced emotion labeling. This underlines the importance of transparency in annotator composition and encourages the use of ``datasheets'' for documenting demographic and modality biases.

\textbf{Potential Societal Impact:} While our dataset is intended for applications in animal welfare, veterinary support, and human-animal interaction research, misuse is possible. For example, automated emotion recognition systems could be deployed in contexts lacking proper ethical oversight, leading to misinterpretation or over-reliance on algorithmic decisions. We encourage future researchers and practitioners to critically assess the contexts of application and to avoid extrapolating beyond the dataset’s scope. Emotion classification, especially in non-verbal species, remains inherently probabilistic and should not be used in isolation for behavioral or legal judgments.

\textbf{Data Transparency and Accessibility:} We plan to release CREMD for academic research with clear documentation, annotation protocols, and guidelines for ethical usage. Participants were informed that their contributions would support open research. All shared data will be de-identified, and access will require agreement to terms prohibiting commercial use or misuse.

In conclusion, while CREMD aims to contribute meaningfully to affective computing and human-animal interaction, we acknowledge the importance of ongoing ethical reflection and community engagement to mitigate risks and support responsible innovation.


\begin{thebibliography}{00}

\bibitem{b1} A. Abzaliev, H. Pérez-Espinosa, and R. Mihalcea, ``Towards dog bark decoding: Leveraging human speech processing for automated bark classification,'' \textit{arXiv preprint arXiv:2404.18739}, 2024.

\bibitem{b2} N. Albuquerque, K. Guo, A. Wilkinson, C. Savalli, E. Otta, and D. Mills, ``Dogs recognize dog and human emotions,'' \textit{Biol. Lett.}, vol. 12, no. 1, p. 20150883, Jan. 2016.

\bibitem{b3} F. Amici, J. Waterman, C. M. Kellermann, K. Karimullah, and J. Bräuer, ``The ability to recognize dog emotions depends on the cultural milieu in which we grow up,'' \textit{Sci. Rep.}, vol. 9, p. 16414, Nov. 2019.

\bibitem{b4} T. Boneh-Shitrit \textit{et al.}, ``Deep learning models for automated classification of dog emotional states from facial expressions,'' \textit{arXiv preprint arXiv:2206.05619}, Jun. 2022.

\bibitem{b5} T. Boneh-Shitrit \textit{et al.}, ``Explainable automated recognition of emotional states from canine facial expressions: The case of positive anticipation and frustration,'' \textit{Sci. Rep.}, vol. 12, art. no. 22611, Dec. 2022.

\bibitem{b6} A. Bremhorst, N. A. Sutter, H. Würbel, D. S. Mills, and S. Riemer, ``Differences in facial expressions during positive anticipation and frustration in dogs awaiting a reward,'' \textit{Sci. Rep.}, vol. 9, art. no. 19312, Dec. 2019.

\bibitem{b7} V. O. Chávez-Guerrero, H. Pérez-Espinosa, M. E. Puga-Nathal, and V. Reyes-Meza, ``Classification of domestic dogs’ emotional behavior using computer vision,'' \textit{Comput. Sist.}, vol. 26, no. 1, pp. 203--219, 2022.

\bibitem{b8} H.-Y. Chen, C.-H. Lin, J.-W. Lai, and Y.-K. Chan, ``Convolutional neural network-based automated system for dog tracking and emotion recognition in video surveillance,'' \textit{Appl. Sci.}, vol. 13, no. 5, art. no. 4596, Apr. 2023.

\bibitem{b9} C. Correia-Caeiro, K. Guo, and D. S. Mills, ``Visual perception of emotion cues in dogs: A critical review of methodologies,'' \textit{Anim. Cogn.}, vol. 26, no. 2, pp. 727--754, Mar. 2023.

\bibitem{b10} E. Csoltova and E. Mehinagic, ``Where do we stand in the domestic dog (\textit{Canis familiaris}) positive-emotion assessment: A state-of-the-art review and future directions,'' \textit{Front. Psychol.}, vol. 11, art. no. 2131, Sep. 2020.

\bibitem{b11} K. Ferres, T. Schloesser, and P. A. Gloor, ``Predicting dog emotions based on posture analysis using DeepLabCut,'' \textit{Future Internet}, vol. 14, no. 4, art. no. 97, Apr. 2022.

\bibitem{b12} V. Franzoni, A. Milani, G. Biondi, and F. Micheli, ``A preliminary work on dog emotion recognition,'' in \textit{Proc. IEEE/WIC/ACM Int. Conf. Web Intell. (WI Companion)}, Thessaloniki, Greece, Oct. 2019, pp. 123--128.

\bibitem{b13} S. Hantke, S. Amiriparian, A. Baird, and B. W. Schuller, ``What is my dog trying to tell me? The automatic recognition of the context and perceived emotion of dog barks,'' in \textit{Proc. IEEE Int. Conf. Multimedia Expo (ICME)}, London, U.K., Jul. 2020, pp. 1--6.

\bibitem{b14} F. Hernández-Luquin, B. Gutierrez-Serafín, H. J. Escalante, H. Pérez-Espinosa, L. Villaseñor-Pineda, and V. Reyes-Meza, ``Dog emotion recognition from images in the wild: DEBIw dataset and first results,'' in \textit{Proc. 9th Int. Conf. Animal-Computer Interaction (ACI)}, Newcastle-upon-Tyne, U.K., Dec. 2022, pp. 1--13.

\bibitem{b15} Z. Kowalczuk, M. Czubenko, and W. Żmuda-Trzebiatowska, ``Categorization of emotions in dog behavior based on the deep neural network,'' \textit{Comput. Intell.}, vol. 38, no. 5, pp. 2116--2133, Oct. 2022.

\bibitem{b16} M. V. Kujala, ``Canine emotions: Guidelines for research,'' \textit{Anim. Sentience}, vol. 14, no. 18, pp. 1--10, 2018.

\bibitem{b17} V. Raudonis, R. Maskeliūnas, and R. Damaševičius, ``Recognition of emotional vocalizations of canine,'' \textit{Acta Acust. united Acust.}, vol. 104, p. 1, Jan. 2018.

\bibitem{b18} D. Mota-Rojas \textit{et al.}, ``Current advances in assessment of dog’s emotions, facial expressions, and their use for clinical recognition of pain,'' \textit{Animals}, vol. 11, no. 11, art. no. 3334, Nov. 2021.

\bibitem{b19} H. Pérez-Espinosa \textit{et al.}, ``Assessment of the emotional state in domestic dogs using a bidimensional model of emotions and a machine learning approach for the analysis of its vocalizations,'' \textit{Res. Comput. Sci.}, vol. 144, pp. 53--65, Oct. 2017.

\bibitem{b20} P. Pongrácz, C. Molnár, and Á. Miklósi, ``Acoustic parameters of dog barks carry emotional information for humans,'' \textit{Appl. Anim. Behav. Sci.}, vol. 100, no. 3–4, pp. 228--240, Jan. 2006.

\bibitem{b21} A. Schirmer, C. S. Seow, and T. B. Penney, ``Humans process dog and human facial affect in similar ways,'' \textit{PLoS One}, vol. 8, no. 9, art. no. e74591, Sep. 2013.

\bibitem{b22} B. M. Waller \textit{et al.}, ``Paedomorphic facial expressions give dogs a selective advantage,'' \textit{PLoS One}, vol. 8, no. 12, art. no. e82686, Dec. 2013.

\end{thebibliography}
\end{document}